\documentclass[letterpaper, 10 pt, conference]{ieeeconf}  

\IEEEoverridecommandlockouts                              

\usepackage{graphicx}
\usepackage{color,soul}
\usepackage{hyperref}
\usepackage{mwe}
\usepackage{subcaption}
\usepackage{amsmath}
\usepackage{tabularx} 
\usepackage{multirow}

\newcolumntype{L}{>{\centering\arraybackslash}m{3cm}}
\usepackage{float}
\usepackage[thinc]{esdiff}
\usepackage{graphics} 
\usepackage[table]{xcolor}
\usepackage{algorithmic}
\usepackage[ruled,norelsize]{algorithm2e} 
\usepackage{caption}
\usepackage{graphics} 
\usepackage{epsfig} 
\usepackage{mathptmx} 
\usepackage{times} 
\usepackage{amsmath} 
\usepackage{amssymb}  
\usepackage[noadjust]{cite}
\usepackage{booktabs}
\usepackage{multirow}
\usepackage{algorithmic}
\usepackage[ruled,norelsize]{algorithm2e} 
\usepackage[noadjust]{cite}
\usepackage{ragged2e}
\newcolumntype{P}[1]{>{\RaggedRight\arraybackslash}p{#1}}

\usepackage{microtype}

\title{\LARGE \bf
Forest Biomass Mapping with Terrestrial Hyperspectral Imaging for Wildfire Risk Monitoring
}

\author{Nathaniel Hanson$^{1,2,*}$, Sarvesh Prajapati$^{1*}$, James Tukpah$^{1}$, Yash Mewada$^{1}$, and Taşkın Padır$^{1,3}$
\thanks{$^{*}$Equal Contribution}
\thanks{Correspondence: {\tt\footnotesize hanson.n@northeastern.edu}} 
\thanks{$^{1}$Institute for Experiential Robotics, Northeastern University, Boston, Massachusetts, USA.}
\thanks{$^{2}$Lincoln Laboratory, Massachusetts Institute of Technology, Lexington, Massachusetts, USA.}
\thanks{$^{3}$Ta\c{s}k{\i}n Pad{\i}r holds concurrent appointments as a Professor of Electrical and Computer Engineering at Northeastern University and as an Amazon Scholar. This paper describes work performed at Northeastern University and is not associated with Amazon.}
}

\begin{document}

\maketitle
\thispagestyle{empty}
\pagestyle{empty}

\begin{abstract}

With the rapid increase in wildfires in the past decade, it has become necessary to detect and predict these disasters to mitigate losses to ecosystems and human lives. In this paper, we present a novel solution --- Hyper-Drive3D --- consisting of snapshot hyperspectral imaging and LiDAR, mounted on an Unmanned Ground Vehicle (UGV) that identifies areas inside forests at risk of becoming fuel for a forest fire. This system enables more accurate classification by analyzing the spectral signatures of forest vegetation. We conducted field trials in a controlled environment simulating forest conditions, yielding valuable insights into the system’s effectiveness. Extensive data collection was also performed in a dense forest across varying environmental conditions and topographies to enhance the system’s predictive capabilities for fire hazards and support a risk-informed, proactive forest management strategy. Additionally, we propose a framework for extracting moisture data from hyperspectral imagery and projecting it into 3D space.

\end{abstract}

\section{INTRODUCTION}

In recent years, there has been a significant increase in the number of wildfires across the globe. Record-setting wildfires have become the norm, with 2020, 2021 and 2023 marking the fourth, third, and first worst years of global wildfires, respectively~\cite{Wildfire2024WF}. These natural disasters cause massive damage to wildlife, property, as well as the atmosphere, and the rise of fires around the world poses a significant problem. As the frequency and intensity of these wildfires increases, the need to accurately monitor risk conditions -- without directly involving forest rangers --- also increases. Remote sensing platforms, airborne and spaceborne, have been used to monitor fire risk with low spatial resolution using electro-optical/infrared (EO/IR) imaging~\cite{veraverbeke2018hyperspectral}. These sensory methods monitor absorbances and reflections of different wavelengths of light to infer the relative abundance of chemical compounds (such as water and chlorophyll)~\cite{shaw2003spectral}. Although remote sensing methods are appropriate to determine fire risk in large geographic areas, understanding fire risk with high spatial resolution is an unsolved problem. This problem is so pressing that the XPRIZE Foundation recently unveiled a grand challenge to encourage the development of autonomous systems in fire detection and mitigation~\cite{xprize}.

\begin{figure}[!t]
\centering
    \includegraphics[width=\linewidth]{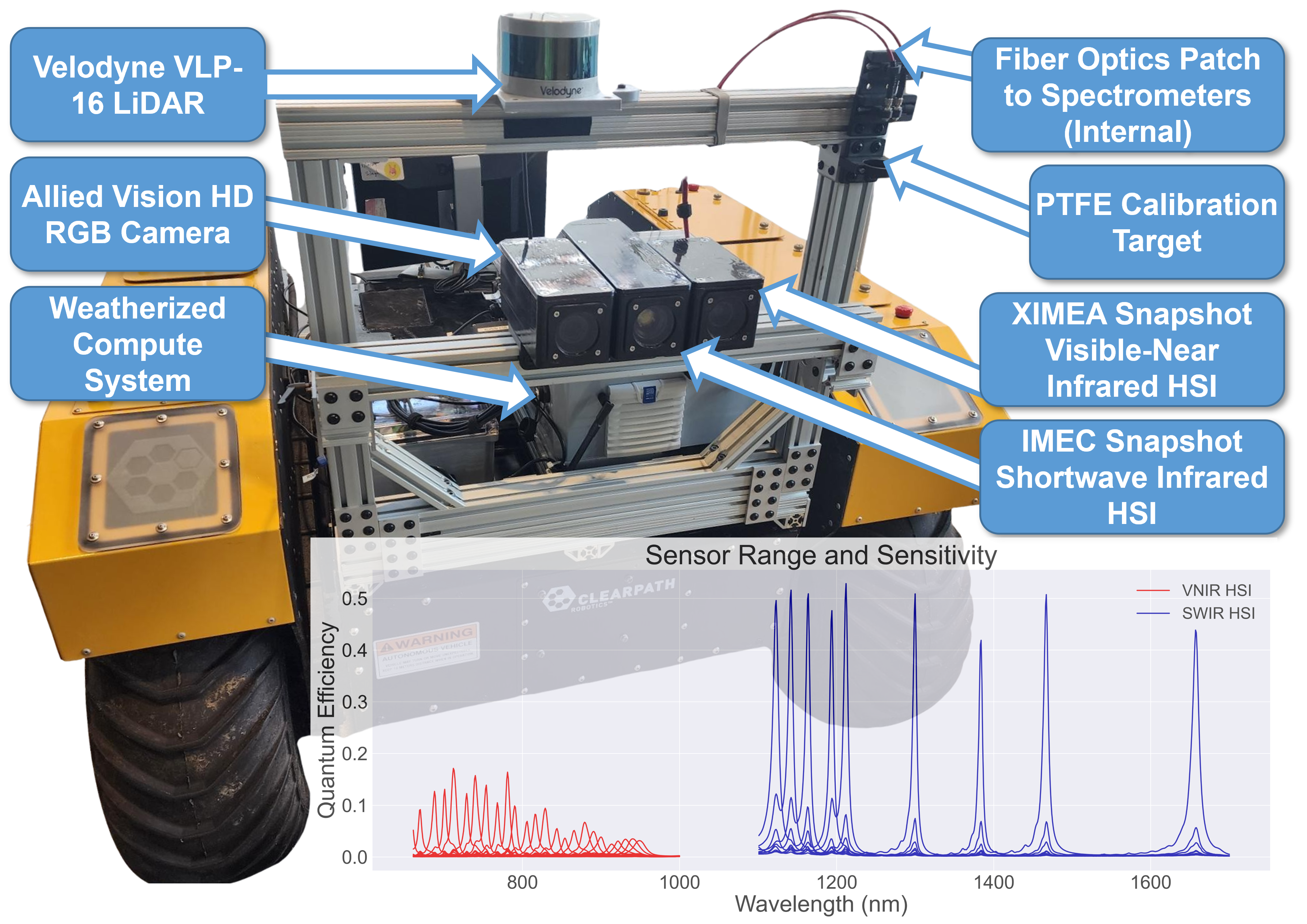}
    \caption{\textbf{Hyper-Drive3D system platform}. The system contains a multi-modal platform for visible-shortwave infrared hyperspectral imaging of natural environments with continuous reflectance calibration provided by solar irradiance measurements. The system is mounted to a Warthog Unmanned Ground Vehicle. (Inset) Quantum efficiency and band spacing of the snapshot imaging array.}
    \label{fig:system_model}
    \vspace{-1.5em}
\end{figure}

In this paper, we introduce a novel system for the risk assessment process in preparation for wildfires through autonomous hyperspectral imaging (HSI). Our system, known as Hyper-Drive3D, grounds remote sensing techniques on a capable unmanned ground vehicle (UGV) that carries a multi-modal imaging system including visible-shortwave infrared hyperspectral cameras, a high-resolution RGB camera, and a 3D LiDAR. Fig.~\ref{fig:system_model} shows the system mounted on the robot base. This multi-modal imaging suite is designed with targeted sensitivity to major spectral absorption features of vegetation, allowing us to assess both biotic and abiotic stresses. As a terrestrial platform, Hyper-Drive3D enables us to investigate the propagation of wildfires and monitor the forest sub-canopy, previously inaccessible to satellites or unmanned aerial vehicles, towards the measurement of biomass present in fine, flash, and heavy fuels~\cite{USDA_Terminology}.

The specific contributions of the full work will include:
\begin{itemize}
    \item Multi-modal platform for assessing forest fire risk in densely vegetated regions.
    \item Pipeline to align, register, and visualize vehicle-centric hyperspectral data and LiDAR data.
    \item Algorithm to quantify regions of highly stressed vegetation using spectral characteristics.
\end{itemize}

\section{RELATED WORK}
While this paper presents a new and novel approach to the prevention and mitigation of wildfires, there has been much work in the literature addressing similar topics; all in response to the increase in wildfires worldwide. Generally, wildfires are uncontrolled fires in a natural ecosystem, including forests, prairies, and croplands. This analysis of prior work examines contemporary approaches by practitioners, analytical methods of wildfire measurement, and the use of autonomous systems in this domain.

\subsection{Practitioner Approaches}
The United States Forest Services defines success in approaching wildfires as ``safely achieving reasonable objectives with the least firefighter exposure necessary''~\cite{usfs}. These objectives balance the need to minimize fatigue and exposure with the risk of further spreading of the fire~\cite{thompson2018wildfire}. Risk management is a significant part of this process that considers values (including loss of life), hazards, and probability~\cite{noonan2021patterns}. Contemporary approaches employ geographic information systems (GIS) to map spatially quantified risk in conjunction with other types of remotely sensed imagery~\cite{chuvieco2010development}. While remote sensing is preferable to cover wide areas and minimize responder exposure, a variety of in-situ measurement approaches yield biomass estimation with a higher accuracy \cite{schunk2016comparison}.

\subsection{Wildfire Quantification}
The fuel moisture content (FMC) \cite{danson2004estimating} focuses on the fire potential of any fuel agent. The measurement takes into account the amount of water contained in a fuel agent (vegetation). Because the reading is based on vegetation characteristics and environmental conditions, it provides more accurate contextual information on fire risk than species or weather conditions. \cite{Argañaraz2018FMC} analyzed pre-fire conditions across a variety of different land covers in the Chaco Serrano subregion in Argentina. They were able to generate FMC maps every 8 days, which accounted for a variety of environmental changes. \cite{Asensio2023FMC} also modeled the dynamic nature of FMC and how it affects the Rate of Spread (ROS) of wildfires.

The use of HSI systems for the classification of wildfires is emerging as an important direction in the quantification of burn areas and drought areas. HSI provides information across a myriad of spectral bands, allowing for more accurate classification based on the spectral profile of vegetation and surface types \cite{vali2020deep}. Spectral indices are simple, explainable mathematical formulas derived from the reflectance or radiance values of HSI wavelengths (or bands) of light. Spectral indices, such as Normalized Difference Vegetation Index (NDVI) \cite{rouse1974monitoring} and Normalized Difference Water Index (NDWI) \cite{gao1996ndwi}, are designed to enhance specific features or properties of the Earth's surface, making it easier to analyze and interpret various environmental and physical conditions. HSI in the shortwave infrared (SWIR) domain also provides the ability to see through smoke and particulates to observe objects not visible with RGB imaging; this is especially relevant for wildfire response \cite{veraverbeke2018hyperspectral}. \cite{Mishra2017HSI} presents the various use cases of hyperspectral cameras to analyze various qualities of plant health. \cite{Thangavel2023HSI} utilizes hyperspectral systems for active fire detection. The combination of preventive and real-time monitoring demonstrates the potential of active HSI to enhance wildfire response.

\subsection{Autonomous Systems}
\cite{Brailey2024Drones} focuses on early detection of wildfires in remote areas by using drones to cover rugged terrain. \cite{Rjoub2022Detection} studies the use of unmanned aerial vehicles (UAVs) with a focus on detecting wildfires through air quality sensors and LiDAR. \cite{Phan2008UAV} uses both UAVs and UGVs to coordinate efforts to identify and mitigate. In our previous work, we used autonomous spectral measuring systems, including both point spectrometers and HSI, to study terrain differences~\cite{hanson2022vast,hanson2023hyperdrive}. A previous version of the Hyper-Drive platform was used to analyze terrain properties such as vegetative health and soil moisture content as priors to traversability estimation~\cite{hanson2024mast}.
\section{TECHNICAL APPROACH}
\begin{figure*}
    \vspace{0.5em}
    \includegraphics[width=0.97\linewidth]
    {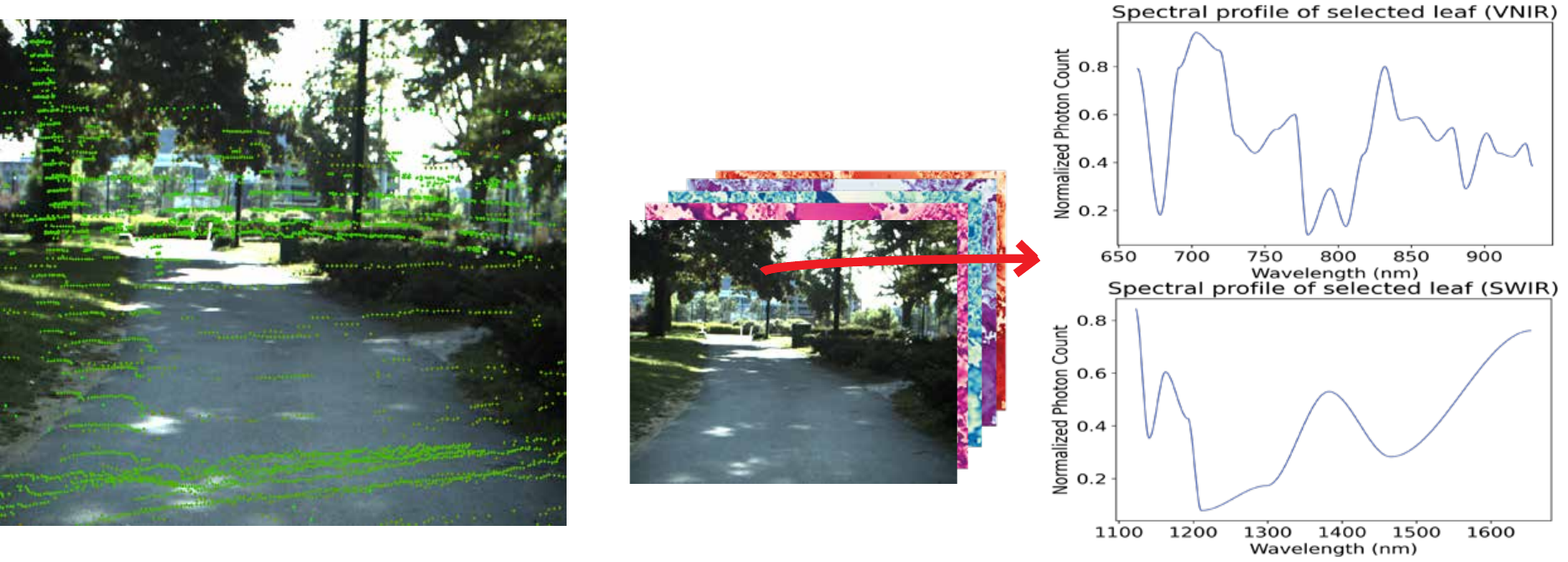}
    \caption{\textbf{Data visualization from field collection.} LiDAR points projected on the RGB image (left), registered hyperspectral cube ($816 \times 684 \times 36$) (center) and spectral profile of a selected pixel from the RGB Image (Top Right: profile from VNIR; Top Left: from SWIR).}
    \label{fig:hsi_collection_viz}
    \vspace{-1.75em}
\end{figure*}
\subsection{Imaging System}
The main sensor design for Hyper-Drive3D is introduced in \cite{hanson2023hyperdrive}. This work overviews the creation of a sensing system that incorporates hyperspectral cameras on a mobile robot. We present the world's first multi-modal dataset containing both visible-short wave infrared hyperspectral images in both urban and off-road environments with a defined labelset building upon open-source terrain label ontologies.

The core vision system consists of two snapshot hyperspectral cameras with complementary wavelength domains. The VNIR camera (XIMEA SNAPSHOT NIR, IMEC) captures light from 660-900 in a 5×5 Fabry–Pérot filter placed in front of a silicon-based photodetector. The SWIR camera (SNAPSHOT SWIR 9 Band, IMEC) captures wavelength information from 1100-1700 nm using a 3×3 Fabry–Pérot filter array on an uncooled Indium Gallium Arsenide (InGaAs) photodetector. Together, this system contains 33 bands of spectral information. The hyperspectral cameras are coaligned with a 5 megapixel RGB machine vision camera with a wide field of view (Alvium 1800 U-507c, Allied Vision). All cameras are contained in an ingress protected housing with chemical resistant glass (Gorilla Glass, Corning), This system provides a high-resolution spatial reference containing the full field of view of the hyperspectral cameras. The inset plot in Fig.~\ref{fig:system_model} shows the efficiencies of the HSI system at each band and the wavelength spacing of the sensor arrays.

The spectra of the environment observed by the HSI systems depend on the quantity and distribution of the available lighting. As this system is designed to operate below the forest canopy, we use a point spectrometer system (Pebble VIS-NIR \& NIR, Ibsen Photonics) to measure observed solar spectra and generate reflectance images using the method from \cite{hanson2024mast}.

Additionally, we have added a 3D LiDAR system (Velodyne VLP-16) to provide additional spatial context and assist in mapping. This addition is also attached to the same camera frame that houses the VNIR, SWIR, and RGB cameras. As an active sensor, the LiDAR does not depend on environmental illumination.

Since the VNIR, SWIR and RGB cameras are mounted next to each other, their imaging axes ($z-axis$) are parallel to each other and translated in the $x-axis$. The VNIR and SWIR images (represented by $\Tilde{x}_v \text{ and } \Tilde{x}_s$) are projected into the frame of the RGB image (represented by $\Tilde{x}_r$). The projection is achieved through the respective homographies of the camera pairing\cite{cvszeliski}.

\begin{equation}
    \Tilde{x}_{rv} \sim \Tilde{\textit{\textbf{H}}}_{rv}\Tilde{x}_v \qquad \text{ and }\qquad \Tilde{x}_{rs} \sim \Tilde{\textit{\textbf{H}}}_{rs}\Tilde{x}_s
    \label{eq:homography}
\end{equation}

Equation~\ref{eq:homography} shows the mapping of VNIR and SWIR camera to RGB camera. Once both cameras are mapped to RGB, due to difference in resolution and translation between the camera, there are some regions which may be blacked out in the projected images. A minimum rectangle is computed in the projected image to ensure the resulting image does not contain patches partial spectral information. Finally all the images are cropped based on the precomputed rectangle and the images are concatenated together to a generate registered cube in the order of RGB, VNIR and SWIR. The dimension of registered cube becomes $w_r\times h_r \times 36$, where $w_r \text{ and } h_r$ is the width and height of raw RGB image respectively.

The LiDAR-Camera extrinsic calibration was performed using a target-based manual method \cite{githubGitHubHeetheshlidar_camera_calibration}. A checkerboard of size 9x14 with each box of size 0.04mm was used as a target for this calibration process; corner points were manually picked for each camera and LiDAR frame. By doing so, we effectively unified all sensors under a single coordinate frame, enabling the projection of data from the 3D LiDAR, RGB, VNIR, and SWIR cameras onto a coherent spatial framework.

\begin{figure}[!b]
    \centering
    \includegraphics[width=0.99\linewidth]{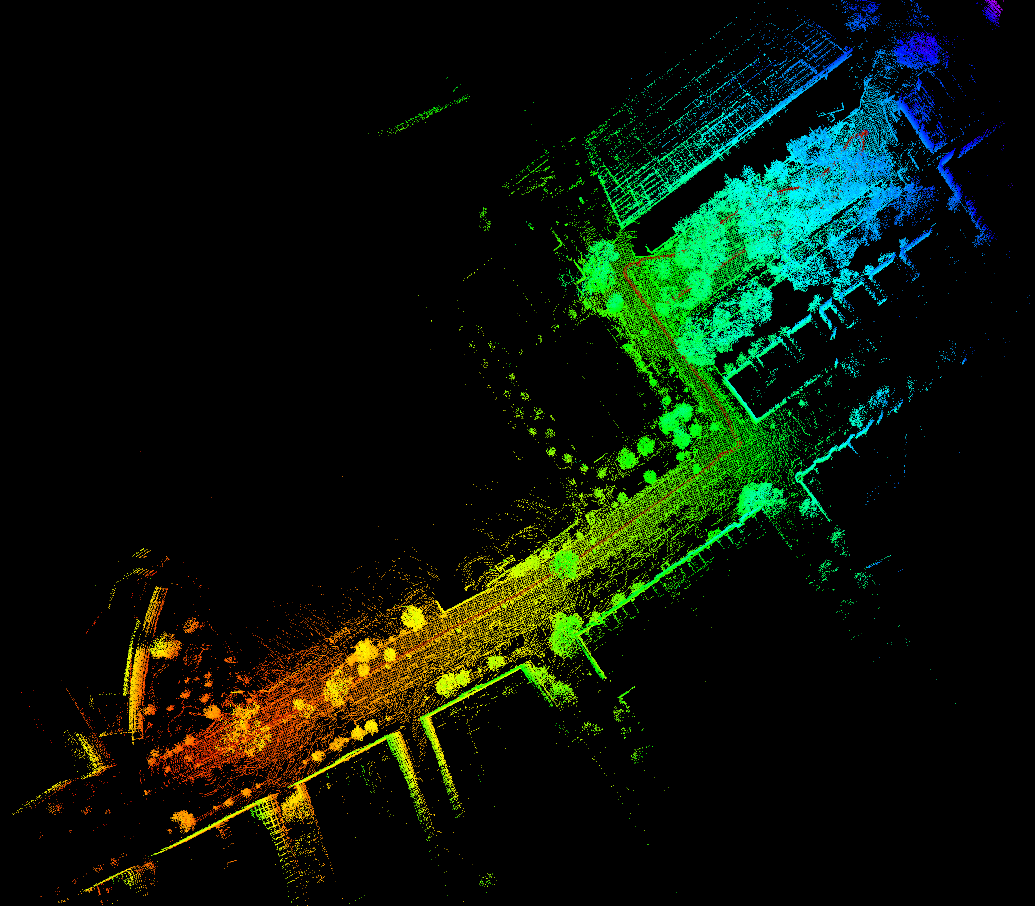}
    \caption{The map illustrates a dense point-cloud map created by LeGO-LOAM \cite{8594299} and shows the trajectory followed by the Warthog Unmanned Ground Vehicle (UGV) during data collection. The point density of the rendered maps shows a sufficient ability to differentiate individual trees and plants.}
    \label{fig:data_collection_route}
\end{figure}
\begin{figure*}
    \centering
    \includegraphics[width=0.3\linewidth]{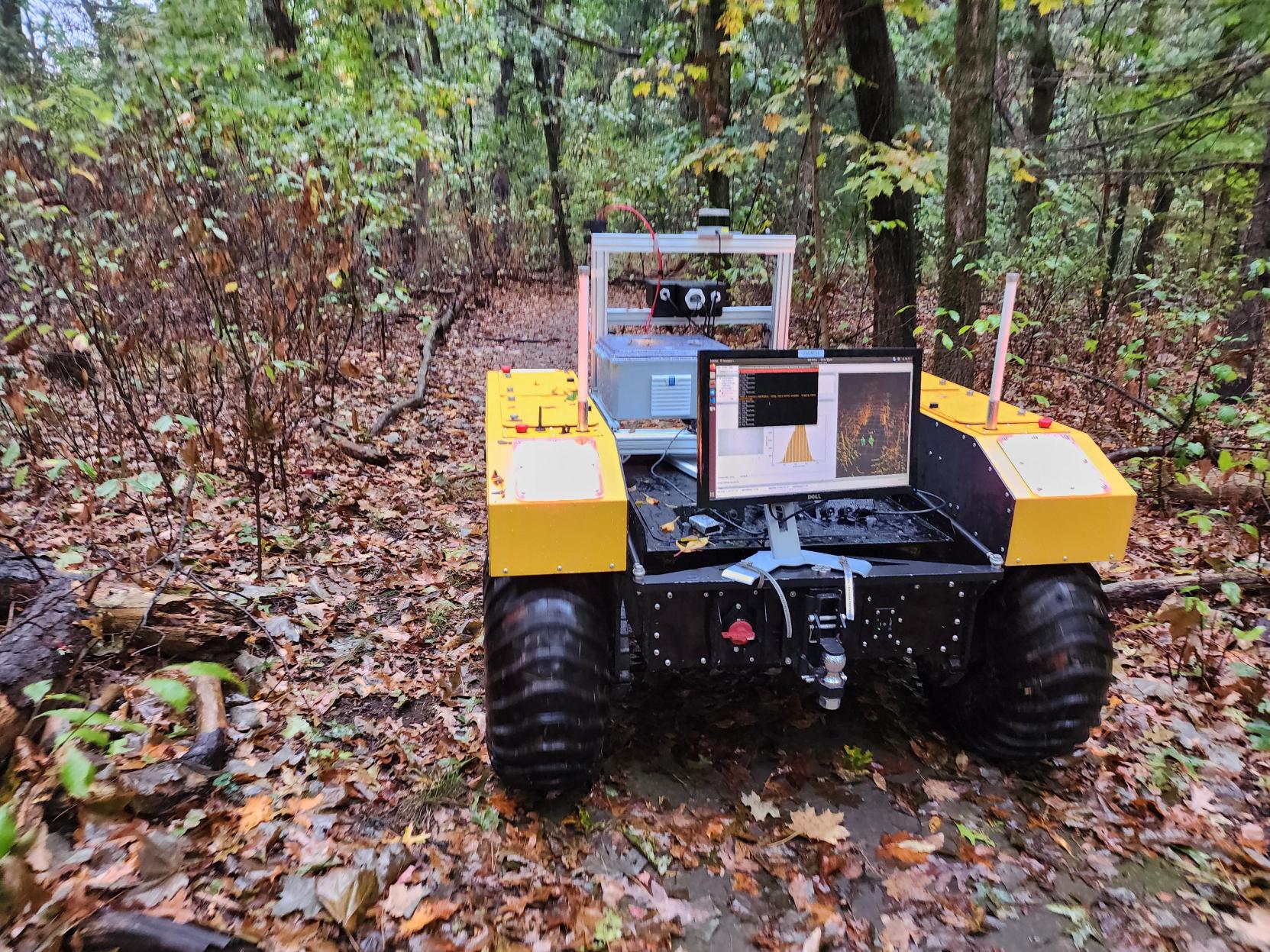}
    \includegraphics[width=0.35\linewidth, height=0.225\linewidth]{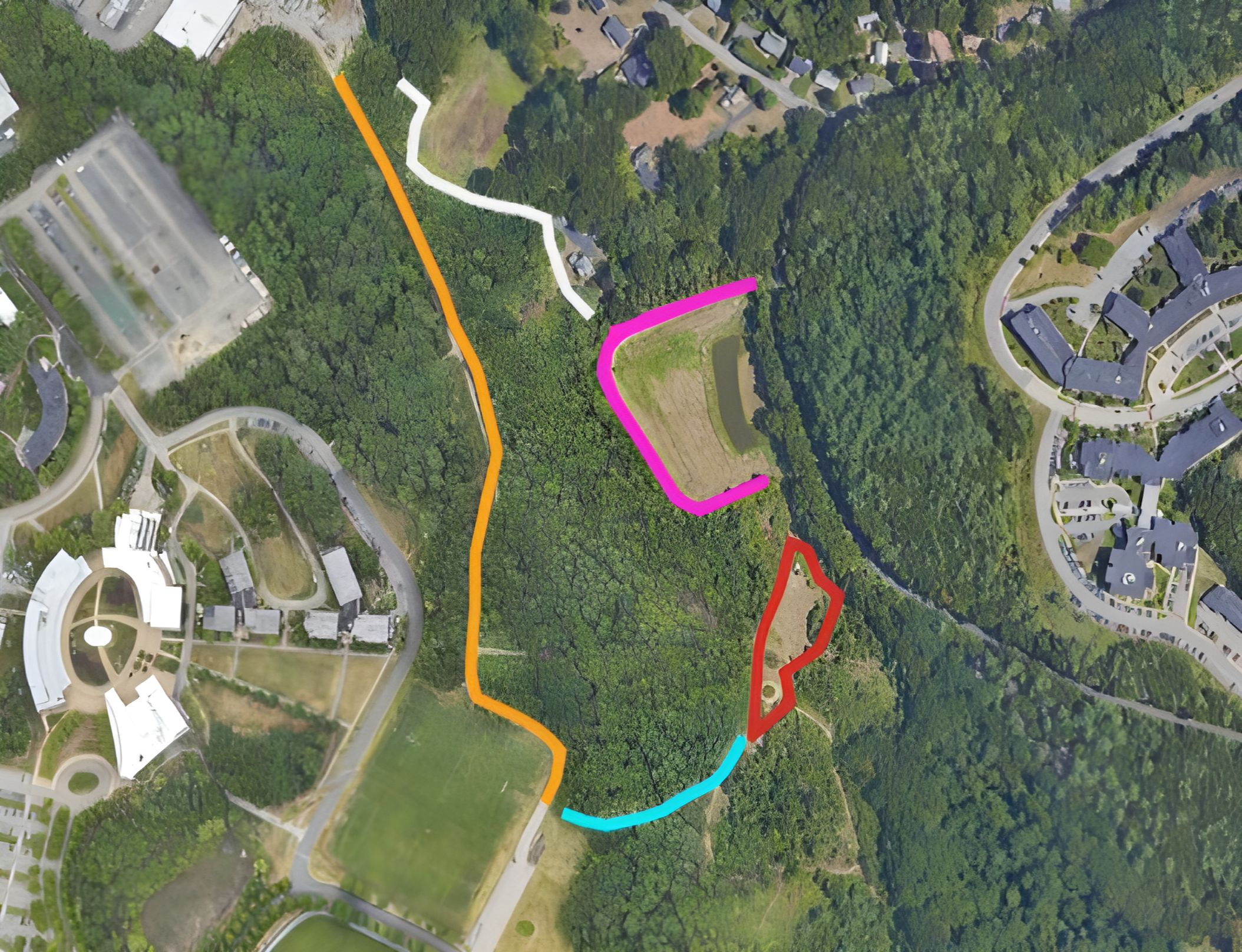}
    \includegraphics[width=0.3\linewidth]{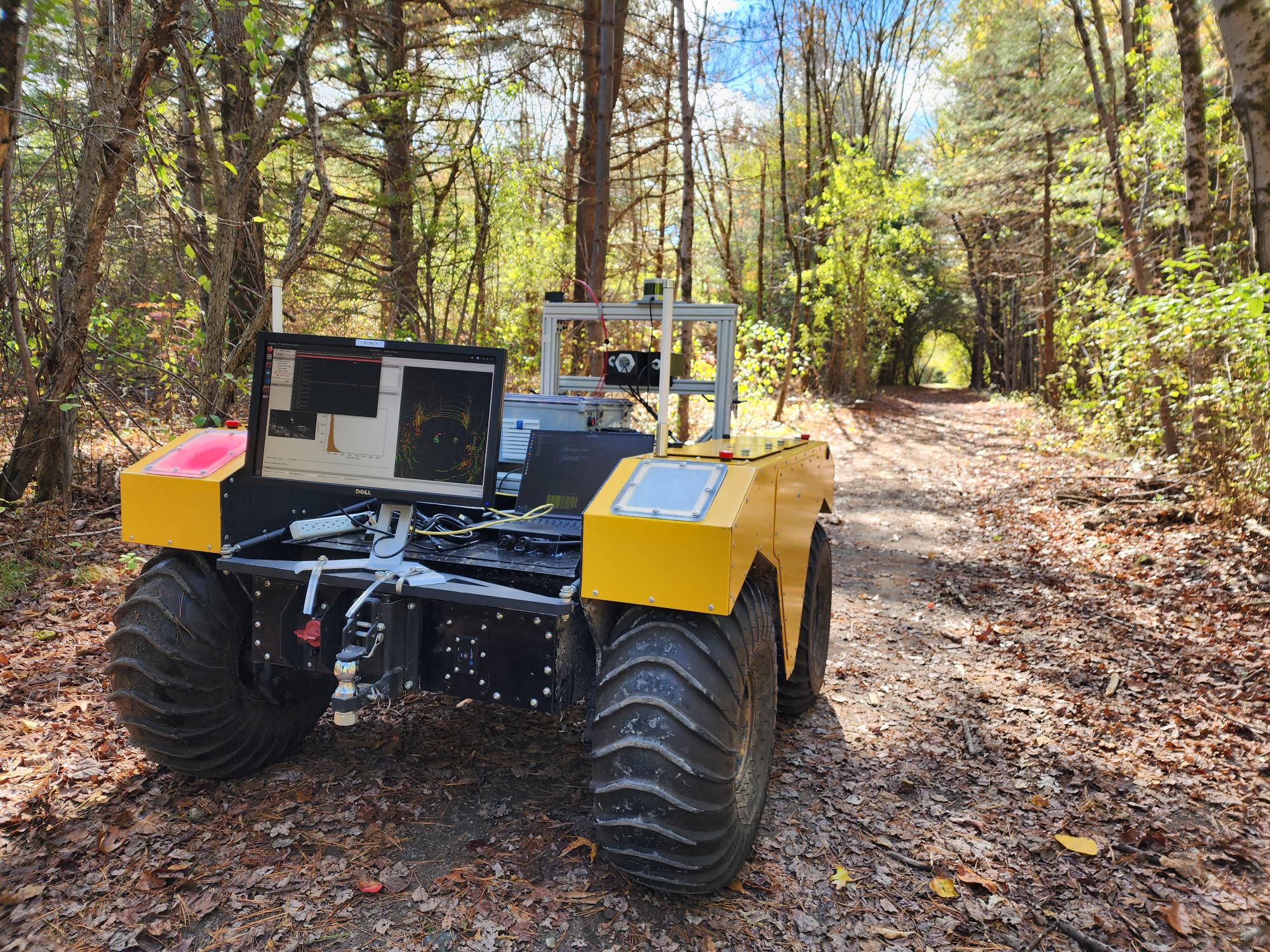}
    \caption{Visualization of all five routes for data collection at Olin (Image courtesy of Google Earth).}
    \label{fig:olin_path}
\end{figure*}
\subsection{Estimating Fuel Moisture Content}
An estimation of the fuel moisture content can be achieved by relying on the combined results of three different spectral indices. Spectral indices are mathematical combinations of different spectral bands that highlight different attributes of materials that are observed via spectral imaging. In the case of the moisture content in vegetation, the three indices that are required are the Normalized Difference Water Index(NDWI), the Normalized Difference Vegetation Index (NDVI), and the Normalized Difference Moisture Index (NDMI). While the NDWI index can provide information on the water content in vegetation it is much more useful for the detection of large water source. The results it provides are limited by sensitivity to background surfaces. This is in part due to its utilization of the Green Spectral band represented by the variable $G$ and the Near Infrared Spectral Band represented by the variable $NIR$ in Equation~\ref{eq:NDWI}. Hence the incorporation of the NDVI which provides information on the health of the vegetation. This is utilized because of the correlation that the healthier the vegetation the more moisture it will retain. Though too has limitations as since it is vegetation specific, the results can not be applied generally across a large section of combined vegetation. This is caused by its utilization of the Red Spectral Band represented by $R$ and the Near Infrared Spectral Band represented by the variable $NIR$ in Equation~\ref{eq:NDVI}. This is the reason for the NDMI index. This specifically focuses on the moisture content of the plant itself, and is derived from the first two indices. Here the utilization of the Shortwave Infrared Band represented by $SWIR$ and the Near Infrared Spectral Band represented by the variable $NIR$ in Equation~\ref{eq:NDMI} allows for a wide application use as well as continuing to provide aspects of vegetation health.
\begin{equation}
    \text{NDWI} = \frac{G - \text{NIR}}{G + \text{NIR}}
    \label{eq:NDWI}
\end{equation}

\begin{equation}
    \text{NDVI} = \frac{\text{NIR} - R}{\text{NIR} + R}
    \label{eq:NDVI}
\end{equation}

\begin{equation}
    \text{NDMI} = \frac{\text{NIR} - \text{SWIR}}{ \text{NIR}+ \text{SWIR}}
    \label{eq:NDMI}
\end{equation}

\section{EXPERIMENTATION AND DATASET}

\subsection{Preliminary Data Collection}

The preliminary data were collected near a field next to Northeastern University's Boston campus, where there were abundant trees and adequate sunlight emulating a sparse forest area. The UGV's high ground clearance and comparatively compact size make it a great robot for operations in forested areas and on unpaved road. As a means of testing the imaging array, the robot was teleoperated along a route with healthy vegetation and under a sparse tree canopy. Fig.~\ref{fig:data_collection_route} shows the path traversed during the collection using LeGO-LOAM \cite{8594299} to post-process the LiDAR point clouds. The data were collected on a sunny day with slight rain on the previous day. The ambient sunlight ranged from full to nebulous cloud cover. The robot was driven on sidewalks and data were collected over a stretch of $\approx{400}\text{ }m$. The collected data were saved as rosbag; LiDAR data was $\approx{10}$ Hz, and hyperspectral data was $\approx{2}$ Hz for storage efficiency. Fig.~\ref{fig:hsi_collection_viz} visualizes the synchronized data, the projection of LiDAR points onto a sampled RGB image, and the sample spectra of a vegetation point in the scene. These experiments confirmed the system's ability to managed multiple sensors with high data rate, on the order of gigabytes per second, while collecting sufficient data to build a dense 3D map upon which we project HSI spectra.

\subsection{Final Data Collection}

Data collection occurred over two separate days, spaced one week apart, at a 180-acre forested area at Olin College, with each day marked by distinct weather conditions. Data was recorded on the first day in rainy weather and on the second day in sunny weather, enabling visual comparisons across varied environmental conditions within the processing pipeline. Each day, five datasets were collected along separate off-road routes, covering a total distance of approximately $1520$ meters. The individual route lengths were $152$ m, $356$ m, $403$ m, $247$ m, and $356$ m for routes 1 through 5, respectively. Fig.~\ref{fig:olin_path} illustrates the data collection routes, with cyan, red, pink, white, and orange representing routes 1 through 5, respectively.
\begin{figure}[h!]
    \centering
    \includegraphics[width=0.95\linewidth, height=0.43\linewidth]{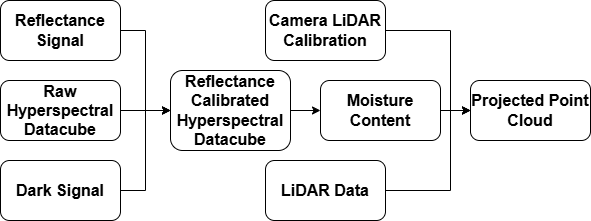}
    \caption{System diagram for generating registered point cloud from Raw Hyperspectral Datacube.}
    \label{fig:system-diagram}
\end{figure}

\begin{figure*}[t!]
    \centering
    \includegraphics[width=0.9\linewidth]{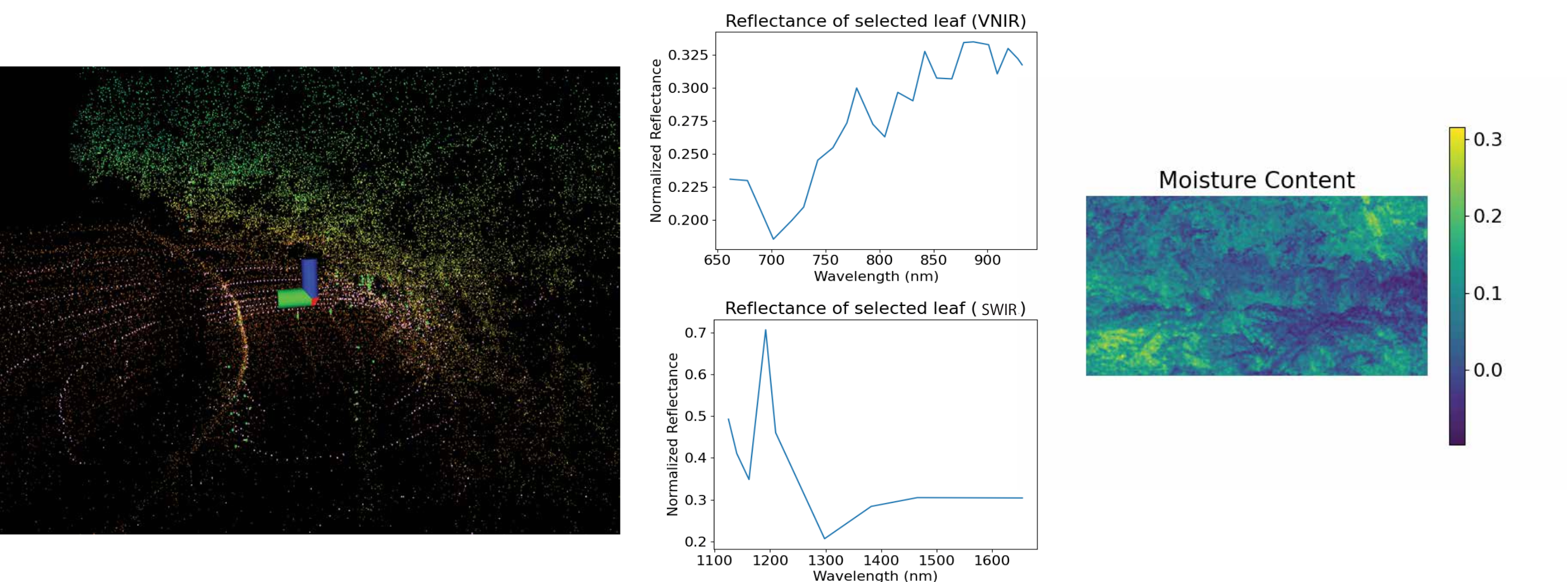}
    \captionsetup{justification=centering}
    \caption{Extracting moisture content from hyperspectral raw cube. Raw Point cloud data shown left is converted to reflectance cube results in the middle. Reflectance is used to calculate moisture content shown right.}
    \label{fig:result}
\end{figure*}
\subsection{Moisture Content}
Figure~\ref{fig:system-diagram} shows the flow for extracting moisture content from the imaging system.
The raw datacube needs to be calibrated to ensure the spectral characteristics are not influenced by lighting conditions~\cite{Klein2008-ue}.
\begin{equation}
    S_{r} = \dfrac{S-S_d}{S_{ref}-S_d}
    \label{eq:reflectance}
\end{equation}
Equation~\ref{eq:reflectance} defines the calculation of the reflectance signal, $S_r$, derived from the raw signal, $S$, using the reference signal, $S_{ref}$, and the dark signal, $S_d$. The reference signal represents uniform reflection from polytetrafluoroethylene (PTFE), selected for its consistent reflection properties across visible and NIR radiation. The dark signal is obtained as the average sensor reading when the device is covered to block any light source from reaching the sensor.

The reflectance-calibrated hyperspectral datacube, $S_r$, can be applied to various tasks~\cite{Cui2022,rs12142335}. In this work, we use specific wavelengths from $S_r$ to determine moisture content within the hyperspectral data. Moisture content is calculated by normalizing the difference between two wavelength channels, 1300 and 1119, as shown in Equation~\ref{eq:moisture}, where $I_M$ is a 2-D array (single-channel image) of moisture content, ranging from $0-1$.
\begin{equation}
    I_M = \dfrac{S_{r1300}-S_{r1119}}{S_{r1300}+S_{r1119}}
    \label{eq:moisture}
\end{equation}

Using this moisture content information, we categorize areas into three risk zones for wildfire susceptibility: high-risk (0-20), medium-risk (20-50), and low-risk (greater than 50). Based on these categories, $I_M$ is converted to $I_{MC}$, representing a categorized RGB image of moisture. This RGB image can be projected onto LiDAR data using the intrinsics and extrinsics from the Camera-LiDAR calibration.

\begin{equation}
{p}_{\text{lidar}} ={T}_{\text{lidar}}^{\text{camera}} \cdot \left( z \cdot {K}^{-1} \cdot {p}_{\text{camera}} \right)
\label{eq:projecting_rgb}
\end{equation}

Equation~\ref{eq:projecting_rgb} shows the projection of 2D image point (${p}_{\text{camera}}$) onto a 3D point cloud (${p}_{\text{lidar}}$) given the depth $z$, camera intrinsics $K$, and transformation between LiDAR and camera $T^{camera}_{lidar}$ which was obtained through camera LiDAR calibration.
\section{Results}

Post data collection the remaining step was to process the data and generate a complete map that provided information about the moisture content projected across the point cloud data captured by the lidar. As mentioned above, the data format the hyper-spectral cameras output is in radiance. This needed to be converted to reflectance so that it could properly be used in the spectral index equations provided in Equations \ref{eq:NDMI} and \ref{eq:moisture}. After successfully converting the spectral radiance into reflectance, the steps remaining were to utilize the equations to receive the moisture value for each individual point in the point cloud, Fig.~\ref{fig:result} shows the extracted moisture content through . That information was then remapped over the point cloud to display the contrasting moisture values discussed in the above section.


\section{Conclusion}
We present a framework for extracting the moisture content of a scene from hyperspectral data and projecting it onto a point cloud. After completing the imaging system and receiving preliminary data from initial tests, we conducted further data collection in early fall under ideal weather conditions to observe changes in vegetation health and forest floor composition. Data collection occurred in a rural, 180-acre forest within the same climate region, an area previously used to construct a hyperspectral dataset in \cite{hanson2023hyperdrive}, albeit without LiDAR information. Data collection took place in two phases over separate days and under varying weather conditions. Using reflectance-calibrated hyperspectral data, we successfully extracted the moisture content of the scene.

\section{Future Work}

The moisture content extracted from hyperspectral data and its projection onto LiDAR can be further processed into 2D cost maps using custom cost functions, which can then be integrated with planning and scheduling algorithms to prioritize and continuously monitor wildfire-prone areas. By integrating sensor data within a coherent spatial framework, each map point (see Fig.~\ref{fig:data_collection_route}) can be linked to its corresponding spectral indices, such as NDVI and NDWI, modeled as random variables. This approach enables the calculation of Conditional Value-at-Risk (CVaR) for these indices, providing a risk or uncertainty metric to gauge regional susceptibility to fire hazards. Combining high-resolution spatial maps with spectral features allows forest rangers to identify areas with elevated biomass accumulation, while also accounting for topographical influences on potential fire spread.


\section*{ACKNOWLEDGMENT}
The authors thank Prof. Kenechukwu Mbanisi from Olin College for coordinating the use of their facilities for data collection. The authors also thank Tom Lutz for his assistance in loading the robot at Northeastern.


\bibliographystyle{IEEEtran} 
\bibliography{references}

\addtolength{\textheight}{-12cm}   

\end{document}